\documentclass[journal]{IEEEtran}
\usepackage{etex}
\usepackage{graphicx}
\usepackage{amsmath}
\usepackage{amssymb}

\usepackage{amsthm}
\usepackage{amsfonts}
\usepackage{url}
\usepackage{color}
\usepackage{epstopdf}
\usepackage{bm}
\usepackage[noend]{algpseudocode}
\usepackage[linesnumbered,ruled]{algorithm2e}

\theoremstyle{definition}

\usepackage{tabu}
\usepackage{subcaption}
\usepackage[font=small]{caption}
\usepackage{multirow}
\usepackage{ltxtable}

\begin{document}

\title{Unsupervised Domain Expansion from Multiple Sources}

\author{Jing~Zhang,~\IEEEmembership{Member,~IEEE,}
        Wanqing~Li,~\IEEEmembership{Senior~Member,~IEEE,}
        Lu~Sheng,~\IEEEmembership{Member,~IEEE,}
        Chang~Tang,~\IEEEmembership{Member,~IEEE,}
        and~Philip~Ogunbona,~\IEEEmembership{Senior~Member,~IEEE}
 \thanks{J. Zhang, L. Sheng are with Beihang University, Beijing, China. W. Li, and P. Ogunbona are with University of Wollongong, Wollongong, Australia. (e-mail: zhang\_jing@buaa.edu.cn; wanqing@uow.edu.au; lsheng@buaa.edu.cn; philipo@uow.edu.au)}
\thanks{Manuscript received XXX; revised XXX.}}

%

\maketitle

\begin{abstract}
Given an existing system learned from previous source domains, it is desirable to adapt the system to new domains without accessing and forgetting all the previous domains in some applications. This problem is known as domain expansion. Unlike traditional domain adaptation in which the target domain is the domain defined by new data, in domain expansion the target domain is formed jointly by the source domains and the new domain (hence, domain expansion) and the label function to be learned must work for the expanded domain. Specifically, this paper presents a method for unsupervised multi-source domain expansion (UMSDE) where only the pre-learned models of the source domains and unlabelled new domain data are available. We propose to use the predicted class probability of the unlabelled data in the new domain produced by different source models to jointly mitigate the biases among domains, exploit the discriminative information in the new domain, and preserve the performance in the source domains. Experimental results on the VLCS, ImageCLEF\_DA and PACS datasets have verified the effectiveness of the proposed method.
\end{abstract}


\IEEEpeerreviewmaketitle

\section{Introduction}

The current machine learning problems have been significantly advanced by deep convolutional neural networks (CNNs)~\cite{Krizhevsky2012}. However, some issues still remain, such as the requirement of large scale labelled training data and catastrophic forgetting. In some applications of light weight systems, such as Robotics and Mobile devices, a system can be pre-trained using data of different environments or users (i.e. different source domains). When the system is utilized in a new environment or by a new user (which are generally from a new domain), it is desirable to adapt the system to the new domain unlabelled data without forgetting all the previous domains. The challenge is that accessing to the source domain data is generally limited due to the storage or privacy issues.

To tackle this problem, this paper proposes a novel unsupervised multi-source domain expansion (UMSDE) method, which is specific to the scenario that data in the source domain are inaccessable. Given the pre-learned source models and the unlabelled new domain data, the goal of UMSDE is to learn an unbiased classifier that is able to classify or recognize real and diverse visual data, i.e. the expanded domain being composed of different domains (including the old source domains and the new domain), rather than merely a small and specific domain. Similar to domain adaptation~\cite{Zhang2019,Shao2015,Ben-David2010,Ganin2016,Long2017}, we assume the source domains and the new domain share a same set of classes while the distributions of data from different domains are different.

\begin{figure}
\begin{center}
\includegraphics[scale=0.35]{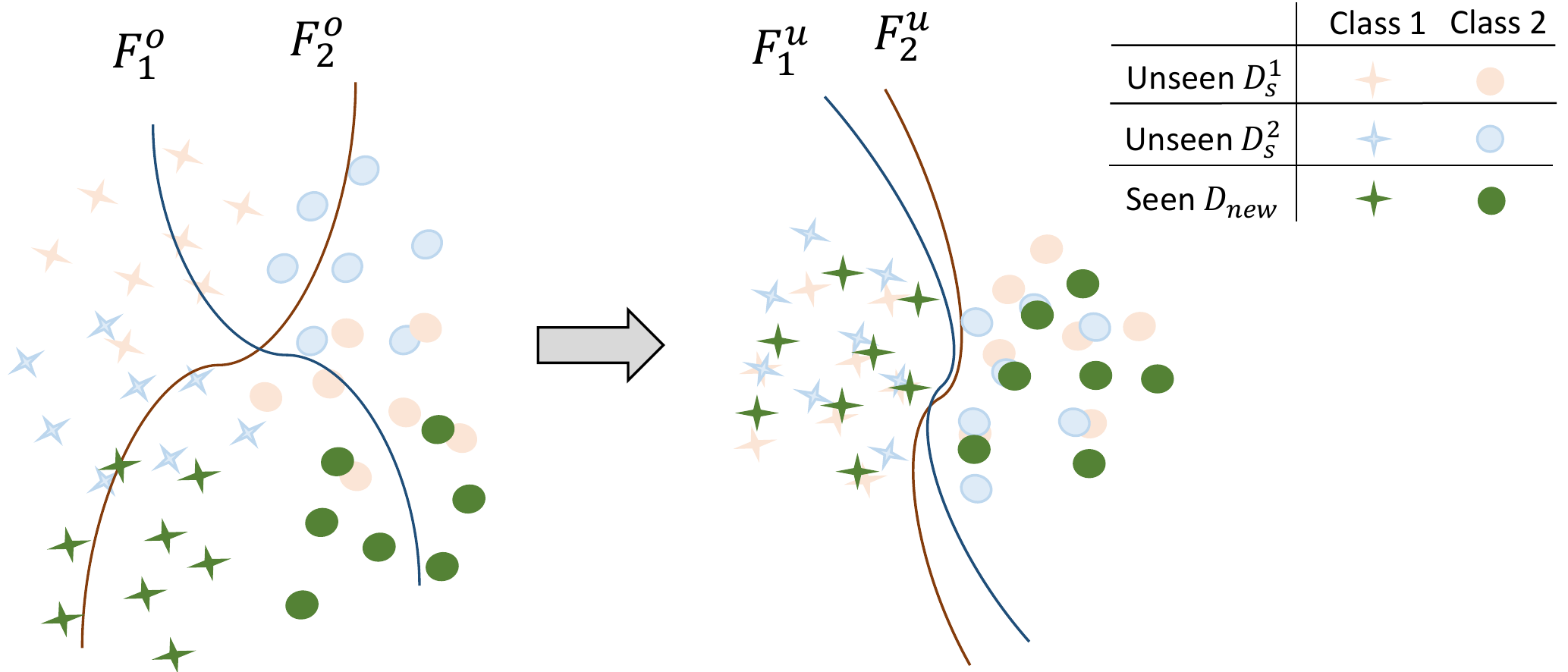}
\caption{The key idea of unsupervised multi-source domain expansion without source data. Given the new domain data and the pre-learned source domain models ($F_1^o,F_2^o$), the updated source models ($F_1^u,F_2^u$) are learned to be unbiased towards the expanded domain (including both the source domains and the new domain). Best viewed in color.}
\label{Fig:alignment}
\end{center}
\vspace{-2em}
\end{figure}

\begin{table*}[!ht]
\begin{center}
\vspace{-1em}
\caption{The comparison between the unsupervised MSDE problem and previous learning schemes. (Notations: $\mathcal{D}_s$: the source domain; $\mathcal{D}_{new}$: the new domain; $\{\mathbf{x}_s,y_s\}_{s=1}^{N_s}$: labelled source domain data; $\{\mathbf{x}_n,y_n\}_{n=1}^{N}$: labelled new domain data; $\{\mathbf{x}_{test},d\}$: the test data with domain label, where $\mathbf{x}$, $y$, $d$ represent data sample, class label, and domain label, respectively.)}
\label{tab:comparison}
\begin{tabular}{|m{4cm}|m{2.8cm}m{1.8cm}m{1.8cm}m{1.8cm}m{1.8cm}|}
\hline
& performance on $\mathcal{D}_s$ & performance on $\mathcal{D}_{new}$ & $\{\mathbf{x}_s,y_s\}_{s=1}^{N_s}$ & $\{\mathbf{x}_n,y_n\}_{n=1}^{N}$ & $\{\mathbf{x}_{test},d\}$\\
\hline
    UDA~\cite{Ben-David2010,Ganin2016,Long2017} & good & good & {\color{red} require $\times$} & not require & - \\  
    UDA w/o $\{\mathbf{x}_s,y_s\}_{s=1}^{N_s}$~\cite{Chidlovskii2016} & {\color{red} no requirement $\times$} & good & not require & not require & - \\  
    DG~\cite{Khosla2012} & {good} & {good} & {\color{red} require $\times$} & not require & not require \\ 
    Incremental Learning~\cite{Li2017d} & good & good & not require & {\color{red} require $\times$} & {\color{red} require $\times$} \\ 
    Supervised DE~\cite{Jung2018} & good & good & not require & {\color{red} require $\times$} & not require \\ 
    Unsupervised MSDE & good & good & not require & not require  & not require \\ 
\hline
\end{tabular}
\end{center}
\vspace{-1em}
\end{table*}

The problem of UMSDE is related to but different from many learning schemes as illustrated in Table~\ref{tab:comparison}.  Thus, the existing approaches to domain adaptation, domain generalization,
incremental learning, and supervised domain expansion are not directly applicable to our UMSDE problem. For example, in unsupervised domain adaptation (UDA), the target domain is defined as the new domain alone, hence, the labelling function to be learned is only required to work on the new domain in DA. In addition, the source domain data are generally available for reducing the distribution shift between different domains. Though a couple of works~\cite{Nelakurthi2018,Chidlovskii2016} propose the domain adaptation methods without the source domain data, they still focus on the performance in the new domain regardless of potential forgetting of the source domain. Another work~\cite{AbdullahJamal2018} proposes a face detector adaptation method without negative transfer or catastrophic forgetting. However, they only concern with the face detection problem and is hard to generalize to other tasks while our method is more generic.
In UMSDE, performance in both source domains and the new domain are equally important. Note that the domain labels of the unseen data are generally unknown. In this case, though the source model can classify the source samples well, the performance on the test data is still unknown because we do not know which domain the test data belong to. In addition, the source domain data are not available. 
The UMSDE is also related to domain generalization. Domain Generalization (DG)~\cite{Khosla2012,Li2017b} uses the data from multiple source domains to learn an unbiased model that can be generalized to the unseen target domain. 
However, DG assumes the data from multiple source domains are available for feasible domain generalization and the new domain data are not exploited when available. 
UMSDE is also different from 
incremental learning~\cite{Li2017d} which requires the labelled training data in the new domain. Jung at al.~\cite{Jung2018} handle supervised domain expansion problem by assuming the labelled data in the new domain are available, which is more closely related to incremental learning~\cite{Li2017d,Lee2017,Zenke2017}, while we focus on the unsupervised setting where only unlabelled data are available for the new domain.

To this end, this paper proposes a method for unsupervised MSDE without accessing to the source domain data. Since the individual source models are learned from the source domains independently, they are inevitably biased toward different source domains. To learn an unbiased model that can perform well on the expanded domain, the biases among different individual domains (including the source domains and the new domain) need to be dealt with. However, due to the unavailable data in the source domains, the biases among different domains are not easy be measured and dealt with explicitly. Therefore, the proposed solution is based on the observation that, after feeding the unlabelled new domain data into different source models, their predicted class probability distributions (pCPDs) over all the classes expose the biases of different domains. Moreover, we find that the entropy of the pCPDs of the new domain data can imply the classification capability of different source models on the new domain data, and thus the degrees of biases of different source domains are roughly discovered and the discriminative information in the new domain is exploited simultaneously. Therefore, this paper proposes to mitigate the domain bias by reducing the difference of the pCPDs of the new domain data in the new domain using different source models, where a mechanism of learning the source model weights based on entropy is properly designed to emphasize more the source domains that are less biased. In addition, in order to transfer the source domain discriminative information and preserve the source domain performance simultaneously, the pCPDs of the new domain data using the original source models and the adapted models are constrained not to change too much.

The main contributions are summarised as follows. 

(1)We introduce a new practical learning problem, named Unsupervised Multi-source Domain Expansion (UMSDE), aims at adapting a system pre-learned on the source domains to the unlabelled new domain without forgetting the source domains and accessing to the source domain data. 

(2) A novel method is proposed for addressing the UMSDE problem using the predicted class probability distributions (pCPDs) of the new domain data on the pre-trained source models. 

(3) Extensive experiments on three real-world cross-domain object recognition datasets and the in-depth analyses verified the effectiveness of the proposed method for the UMSDE problem.

\begin{figure*}
\begin{center}
\includegraphics[scale=0.52]{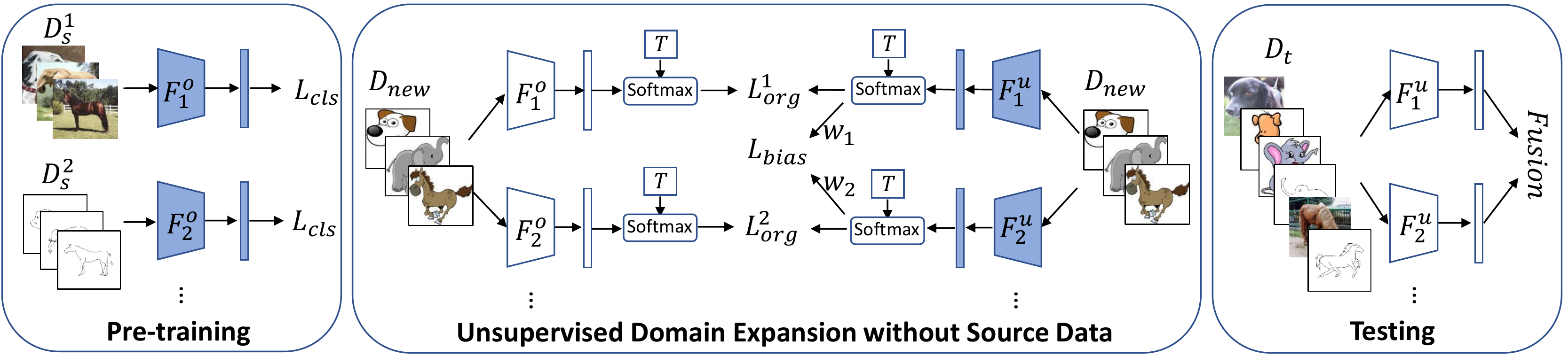}
\end{center}
\vspace{-1em}
\caption{An overview of the proposed UMSDE method. The original source models ($F_1^o,F_2^o,\cdots$) are pre-trained on labelled data from different source domains. Next, to learn unbiased models that perform well on the expanded domain without the source domain data, the source models are updated by feeding the new domain data into the source models, where the biases are mitigated by a weighted loss between the outputs of pairs of updated models and $F_1^o,F_2^o,\cdots$ are used to initialize as well as to constrain the updated source models. The updated source models ($F_1^u,F_2^u,\cdots$) are fused for testing on the expanded target domain (including both the source domains and the new domain). Unshaded blocks indicate fixed network parameters.}
\label{Fig:framework}
\end{figure*}

\section{Related Work}
UMSDE is a new problem with unique requirements that have not been explicitly addressed in the domain adaptation (DA) and domain generalization (DG), and incremental learning literature.
Little study on UMSDE has been reported in the past. In this section, works on DA and DG, as well as incremental learning are briefly reviewed since the former shares the aspect of the problem to be addressed that both need to reduce biases between domains and the later share the common aspect to the problem addressed in this paper both do not access the previous data.

\subsection{Domain Adaptation and Generalization}
There has been extensive research on domain adaptation, as surveyed in~\cite{Pan2010,Zhang2019}. The domain shift is generally reduced by matching the statistic moments~\cite{Pan2011,Zhang2017,Long2015a}, or by relying on the domain adversarial nets~\cite{Ganin2016,Tzeng2017,Saito2018}. For example, \cite{Long2015a} propose a Deep Adaptation Network (DAN) architecture, where the mean embeddings of different domain distributions are explicitly matched in a reproducing kernel Hilbert space. Differently, \cite{Ganin2016} enforce the learned features to be indiscriminate with respect to different domains using an adversarial loss. Some methods focus on multi-source domain adaptation (MSDA)~\cite{Sun2015a,Erfani2017,Xu2018} using data from multiple source domains for more effective transfer. Two main approaches are identified for MSDA. The first commonly used approach is to train a classifier per source and combine these multiple base classifiers by assuming that the distribution of the target domain is a mixture of the multiple source distributions~\cite{Xu2018}. Another commonly used approach is to use some feature mapping methods to learn the domain invariant features~\cite{Erfani2017}. However, both the DA and MSDA methods require the availability of data from both the source domains and the new domain, and only concern with the performance on the new domain. A couple of works~\cite{Nelakurthi2018,Chidlovskii2016} propose the domain adaptation methods without the source domain data. However, some of them require labelled data in the new domain~\cite{Nelakurthi2018} and these methods are not scalable to the deep CNN-based models. In addition, they still focus on the performance in the new domain regardless of potential forgetting.

Domain Generalization (DG)~\cite{Blanchard2011,Khosla2012,Muandet2013,Xu2014,Li2017b} uses the data from multiple source domains to learn an unbiased model that can be generalized to the unseen target domain. Most of the existing work tackle this problem by learning domain invariant and compact representation from multiple source domains~\cite{Blanchard2011,Muandet2013}, or by learning robust classifiers from multiple sources to generalize well on unseen target domain~\cite{Khosla2012,Xu2014,Li2017b}. However, DG assumes the data from multiple source domains are available for feasible domain generalization and the new domain data are not exploited when available. 

\subsection{Incremental Learning}
The incremental learning~\cite{Lee2017,Zenke2017,Li2017d} paradigm tries to learn incrementally without access to the source data. The key challenge in such a scenario is how to alleviate catastrophic forgetting, a major drawback in neural network-based models when being incrementally learned through new data. However, labelled data from the new domain are required and the test sample is generally accompanied by the information of which domain it belongs to~\cite{Li2017d}. By contrast, in our UMSDE problem, the new domain data are assumed to be unlabelled and no prior knowledge on which domain a test sample belongs to is required.

\section{UMSDE without Source Data}
This section presents the proposed method in details. We are given $m$ source models pre-learned using data from a set of $m$ different source domains $\mathcal{D}_s = \{\mathcal{D}_s^{1}, \cdots, \mathcal{D}_s^{m}\}$, where the $i^{th}$ original source model pre-learned on the $i^{th}$ source domain $\mathcal{D}_s^{i}$ is denoted as $F(\mathbf{x};\theta_i^{(o)})$ (or $F_i^{o}$ for short) which is parametrised by the parameters $\theta_i^{(o)}$. In addition, a new domain $\mathcal{D}_{new} = \{\mathbf{x}_n\}_{n=1}^{N}$ of $N$ unlabelled data drawn from distribution $P_{new}(\mathbf{x})$ is available. The objective of unsupervised UMSDE is to learn an unbiased model which performs well on the test data of the expanded target domain $\mathcal{D}_{t} = \mathcal{D}_s\cup\mathcal{D}_{new}$. Due to the domain shift, the source domains and the new domain have different distributions. Moreover, the degree of domain biases of the $m$ sources, $\mathcal{D}_s^1, \cdots, \mathcal{D}_s^m$ are different.
Effective handling of the biases is a key in the UMSDE without source data. 

\subsection{Unbiased Model Learning}
To learn a model that performs well on the expanded domain, the biases of the source models should be alleviated. A possible solution is to reduce the distribution divergence among the source domains as well as the new domain. However, since the source domain data are inaccessible, the distribution divergence is not able to be measured and reduced explicitly. Hence, we propose to reduce the biases by using the predicted class probability distributions (pCPDs) of the new domain data produced by different source models.

We take the case of two source domains as an example (as shown in Figure~\ref{Fig:alignment}). The intuition is that if the two source domains have a large distribution divergence, the learned individual source models are very different from each other. When a new domain sample is fed into the two source models, the pCPDs produced by them will be very different, which expose the shift among different source domains. 
More importantly, motivated by~\cite{Saito2018}, we find that this difference can also reveal the samples from the new domain that are not in the support of the intersection of the two source domains, since if a new sample is within the support of the intersection of the two source domains then the pCPDs of this sample produced by the two source models should be similar.

Therefore, we propose to reduce the divergence using the $L_2-distance$ among the soft pCPDs of the new domain data produced by different source models. Hence, the domain alignment loss for reducing the biases of the $i^{th}$ source model is defined as follows:
\begin{equation}
\small
\hspace{-1em}
\mathcal{L}_{bias}^i\!=\!\frac{1}{N} \sum_{n=1}^{N}\!\sum_{j=1, j\neq i}^m  \|\sigma(o(\mathbf{x}_n;\theta_i^{(u)})/T)-\sigma(o(\mathbf{x}_n;\theta_{j}^{(u)})/T)\|_2^2
\label{eqt:shift}
\end{equation}
where $\theta_i^{(u)}$ represents the network parameters of the $i^{th}$ updated source model (which is initialized by the parameters of the $i^{th}$ original source model $\theta_i^{(o)}$),
and $\sigma(o(\mathbf{x}_n;\theta_i^{(u)}))$ is the $C$-dimensional pCPD produced by the $i^{th}$ source model $F(\mathbf{x};\theta_i^{(u)})$ using the softmax function for the new domain sample $\mathbf{x}_n$, 
$o(\mathbf{x}_n;\theta_i)$ is a $C$-dimensional vector of logits predicted by the $i^{th}$ source model, $C$ is the total number of classes, $T$ is a is a hyperparameter representing the temperature value (which is normally set to 1 in general classification tasks). The use of temperature here is inspired by the knowledge distillation~\cite{Hinton2014}. 
The motivation is that the obtained pCPDs of the new domain data by the source models may not be able to well imply which class the new domain sample belongs to due to the domain bias, but they can indicate the information of class relationships (e.g. which classes the target sample is more close to). If $T$ is small, the class relationships are not able to be effectively revealed since each target sample is encouraged to be very similar to one specific source class. If $T$ is large, the probability distribution over classes is softer, which gives a richer representation of the target sample. However, if the $T$ is too large, the class relationship information will be destroyed since the probabilities of all the classes are similar. Hence, a proper value of $T$ will help the preservation of target data information.

\subsection{Source Model Importance Weight Learning}
\begin{figure}
\includegraphics[scale=0.23]{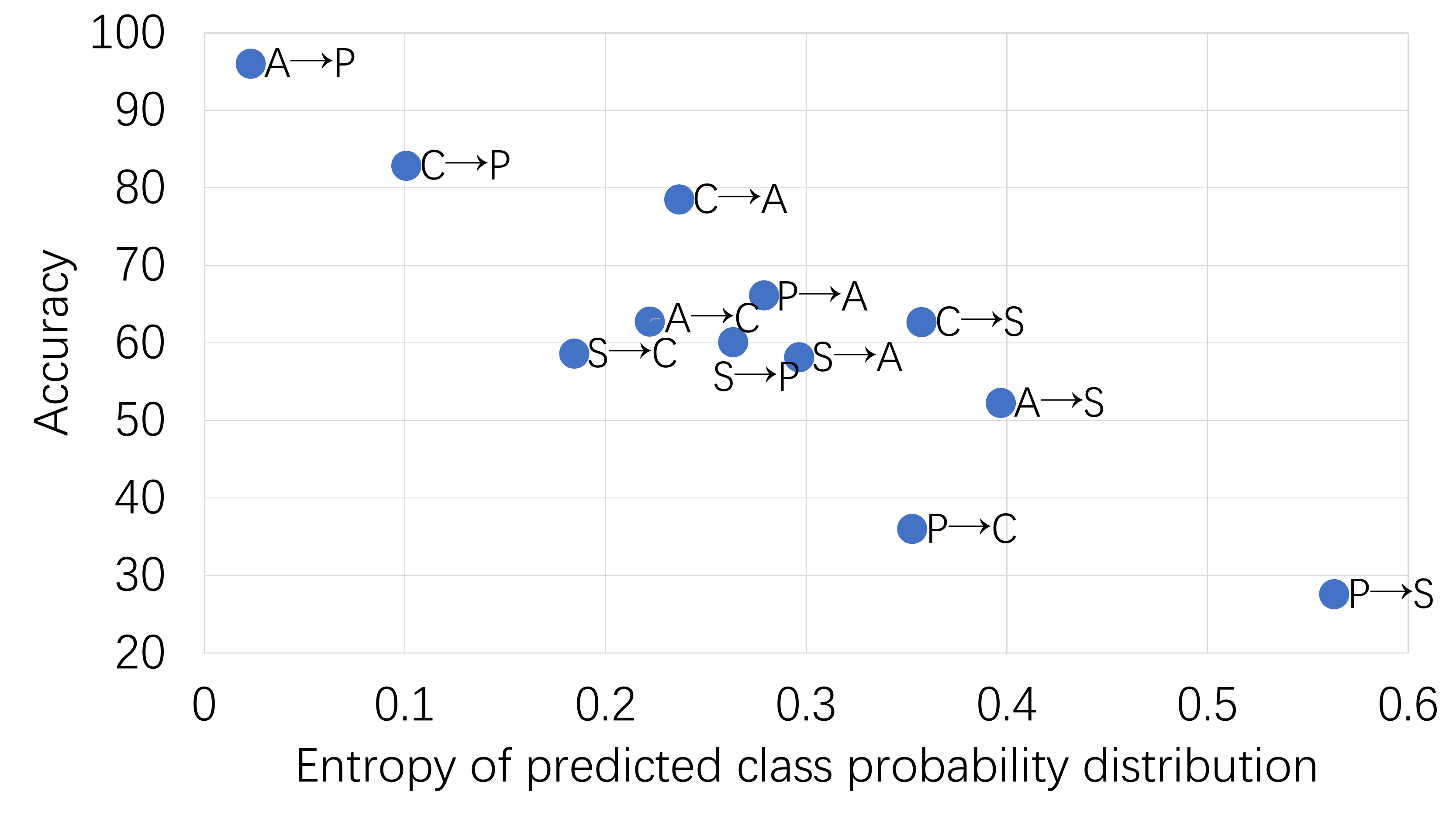}
\vspace{-1em}
\caption{The relationship between accuracy and the mean entropy of predicted class probability distribution on the cross domain recognition tasks of PACS dataset (which contains 4 different domains forming 12 cross domain recognition tasks). They are almost linearly negatively related.}
\label{fig:entropy}
\end{figure}

As mentioned, different source domains are biased to different degrees. Hence, of a source domain is less biased than others, and the corresponding source model should not be changed too much when reducing the biases. 
We propose to use the entropy of the pCPDs of the new domain data produced by different source models to measure the degrees of biases and exploit more discriminative information in the new domain simultaneously. 
Intuitively, if the entropy is low, the decision boundary produced by the source model lies in low density areas~\cite{Grandvalet2005} of the new domain data (i.e. the source model can successfully classify the new domain data). On the other hand, if the source model performs poorly on the new domain, most of the new domain samples will likely produce vague class probability values on many classes rather than a large value on a certain class and thus the entropy of the pCPDs of the new domain data is high. We also empirically visualize the relationship between the classification accuracy and the entropy of the pCPDs in Figure~\ref{fig:entropy}, which is almost linearly negatively related.
Therefore, a source model with better (poorer) classification capability on the new domain data may be less (more) biased and can discover more (less) discriminative information in the new domain, and thus a smaller (larger) importance weight should be assigned to the bias loss in Eq.~\ref{eqt:shift} when updating the source model. Thus, a new weighting scheme based on the entropy of pCPDs is proposed as follows,

\begin{equation}
w_i = \frac{\exp (E_i/T_0)}{ \sum_{j=1}^m \exp (E_j/T_0)}
\label{eqt:weights}
\end{equation}
where  
$E_i\!\!=\!\!\!\frac{1}{N}\!\sum_{n=1}^{N}\!H(\sigma(o(\mathbf{x}_n;\theta_i^{(u)})))$, $H(\cdot)$ is the information entropy function, $T_0$ is the temperature to control the weight differences among sources (i.e. similar to $T$, a higher temperature value gives a softer probability distribution over all the weights). The weights $w_i$ are assigned to the $i^{th}$ source model when reducing the domain bias using Eq.~\ref{eqt:shift}. 

Previous works also learn the source model weights for multi-source domain adaptation (MSDA)~\cite{Mansour2009,Duan2012c,Xu2018}. However, the learning of the importance weights in previous works generally rely on data from both source domains and the new domain, while only the source models and the unlabelled new domain data are required in our method. In addition, the criteria of the learning of the weights are different. In previous MSDA methods, the source model weights are generally learned based on the domain bias between each source domain and the new domain. By contrast, our source model weights are learned based on the classification capabilities of each source model on the new domain data. Thus, the proposed method not only considers the degree of domain bias between each source domain and the new domain but also considers the classification capability (which cannot be implied merely through the domain bias) of each source model, which benefits to the performance of the new domain and the overall UMSDE tasks. Moreover, the use of the source weights are also different from the previous MSDA work. The source model weights in previous MSDA methods are generally directly used for constructing the new target domain classifier (i.e. the weighted combination of multiple source classifiers). By contrast, our source model weights are used for reducing the bias among different domains.

\subsection{Source Domain Information Preservation}
If we simply reduce the divergence among the pCPDs by different source models, the source domain discriminative information is missing and performance on the original source data will be destroyed. In addition, some trivial solutions will be obtained without any constraints on $\mathcal{L}_{bias}$. Hence, we propose another loss that after adaptation, the pCPDs of the new domain data produced by the updated source models should not be far away from that of the original source models, which is similar to learning without forgetting~\cite{Li2017d}. This loss not only preserves the discriminative information and source domain performance, but also acts as a regularization to avoid trivial solution when reducing the domain bias. The loss $\mathcal{L}_{org}^i$ for the $i^{th}$ source model is defined as follows,
\begin{equation}
\small
\mathcal{L}_{org}^i =  \min_{\theta_i} \frac{1}{N} \sum_{n=1}^{N} \|\sigma(o(\mathbf{x}_n;\theta_i^{(u)})/T)-\sigma(o(\mathbf{x}_n;\theta_{i}^{(o)})/T)\|_2^2
\label{eqt:original}
\end{equation}
For simplicity, the temperature ($T$) here is set identically to that in Eq.~\ref{eqt:shift}

\subsection{Overall Model}
By combining the multi-source alignment loss (Eq.~\ref{eqt:shift}) with the weights defined in (Eq.~\ref{eqt:weights}) and source information preservation loss (Eq.~\ref{eqt:original}), the overall objective
for the $i^{th}$ source model is defined as follows,
\begin{equation}
\mathcal{L}_{overall}^i = \mathcal{L}_{org}^i + \lambda\cdot w_i\cdot\mathcal{L}_{bias}^i
\label{eqt:overall}
\vspace{-0.3em}
\end{equation}
where $\lambda$ is a trade-off parameter. 
Hence, the updated source models are less biased and the discriminative information is preserved.

The source models are updated one by one in each step, which means when one model is updated the rest of the source models are kept fixed. 
After learning, the updated and/or original source models can be fused to construct a model for the expanded domain. In this paper, the model for the expanded domain is constructed in an ad-hoc way. First, the final classification scores $y_{final}(x_t)$ of a test sample $x_t$ is the average score fusion over all the updated models, i.e.
\begin{equation}
\small
y_{final}(x_t) = \frac{1}{m} \sum_{i=1}^m \sigma(o(\mathbf{x}_t;\theta_i^{(u)}))
\label{eqt:fusion1}
\vspace{-0.5em}
\end{equation}
Thus, the final model leverages the discriminative information from all the updated source models with less biases. A summary of the proposed method is provided in Figure~\ref{Fig:framework}.

If the original source models could be kept as well, the $c{th}$ dimension of the final classification scores ($y_{final}^c(x_t)$) of a test sample $x_t$ is the average fusion of the max fusion of each original and updated source model, i.e.
\begin{equation}
\small
y_{final}^c(x_t) = \sum_{i=1}^m(\max(\sigma^c(o(\mathbf{x}_t;\theta_i^{(u)})),\sigma^c(o(\mathbf{x}_t;\theta_i^{(o)}))))
\label{eqt:fusion2}
\end{equation}
where $\sigma^c(o(\mathbf{x}_t;\theta_i))$ is the probability of $\mathbf{x}_t$ belonging to class $c$, and $[y_{final}^1(x_t),...,y_{final}^C(x_t)]^T$ forms the final classification scores of $x_t$. The max fusion selects the individual model with more confidence on a certain class and the average fusion over different individual models leverage the information from different source models. 
\section{Experiments}
This section presents the experiments for evaluating the proposed method.
The experiments were conducted on three cross-domain object recognition datasets: VLCS~\cite{Torralba2011,Khosla2012}, ImageCLEF\_DA\footnote{\url{http://imageclef.org/2014/adaptation}}, and PACS~\cite{Li2017b}. Each dataset contains multiple domains.
The VLCS dataset consists of images from PASCAL VOC2007 (V)~\cite{Everingham}, LabelMe (L)~\cite{Russell2008}, Caltech-101 (C)~\cite{Fei-Fei2006}, and SUN09 (S)~\cite{Choi2010} datasets, each of which represents one domain. Five categories are shared among these datasets.
The ImageCLEF\_DA is a benchmark dataset for the ImageCLEF 2014 domain adaptation challenge. Twelve shared categories are selected from four public image datasets: Caltech-256 (C)~\cite{Griffin2007}, ImageNet ILSVRC2012 (I)~\cite{Deng2009}, PASCAL VOC2012 (P)\cite{Everingham}, and Bing (B)~\cite{Bergamo2010}, forming four different domains.
The PACS dataset is a recently collected cross-domain recognition dataset. 
It consists of four different domains (Art painting (A), Cartoon (C), Photo (P), Sketch (S)) with seven common categories. The dataset is created by combining shared classes from four image datasets: Caltech256 (Photo)~\cite{Griffin2007}, Sketchy (Photo, Sketch)~\cite{Sangkloy2016}, TU-Berlin (Sketch)~\cite{Eitz2012} and Google Images (Art painting, Cartoon, Photo). 
\subsection{Implementation Details} 
All three datasets have four different domains. In the experiments, three domains were used as source domains and the rest was used as the new domain, 
resulting four different cases from each dataset. Samples in each domain are randomly divided into a training set (70\%) and a test set (30\%). The training data of the source domains were used for training the source models and then will not be used in the UMSDE task. The pre-trained source models and the unlabelled training data of the new domain were used for training the proposed method and the unseen test data were used for evaluating the learned model.

The individual deep source models were obtained separately by fine-tuning an AlexNet~\cite{Krizhevsky2012} for the VLCS dataset, a ResNet50~\cite{He2016} for both ImageCLEF\_DA and PACS datasets using Caffe~\cite{Jia2014a}, all networks were initialized by a corresponding model pre-trained on ImageNet. The proposed UMSDE without source data method was implemented using PyTorch\footnote{\url{http://pytorch.org/}}. The trade-off parameter was set to $\lambda=10$ and the temperatures for aligning different domains and for the source model weights were set to $T=3$ and  $T_0=0.1$ respectively. The learning rate was 1e-6. Note that these parameters were selected empirically. Since the proposed method addresses a novel problem in UMSDE, fair comparisons cannot be made with previous methods.

\begin{table}[!ht]
\begin{scriptsize}
\begin{center}
\caption{The accuracies (\%) on the VLCS dataset.}
\label{tab:VLCSUMSDE}
\vspace{-1em}
\hspace{-1em}
\begin{subtable}{.492\linewidth}
\caption{\scriptsize Source domains: L, S, V.}
\vspace{-0.5em}
\begin{tabular}{|m{0.77cm}|m{0.25cm}m{0.75cm}m{0.86cm}|}
\hline
Domain & Base & M1 & M2 \\
\hline
    C(New) & 92.92 & \textbf{96.18}$\pm$0.1 & 94.86$\pm$0.5  \\  
    L($S_0$) & \textbf{64.87} & 62.65$\pm$0.2 & 64.09$\pm$0 \\  
    S($S_1$) & \textbf{77.64} & 75.82$\pm$0.3 & 77.35$\pm$0.5 \\  
    V($S_2$) & \textbf{76.01} & 75.72$\pm$0.1 & 75.97$\pm$0.4 \\  
    Expanded & 77.86 & 77.59 & \textbf{78.07} \\
\hline
\end{tabular}
\end{subtable}
\hspace{0.5em}
\begin{subtable}{.492\linewidth}
\caption{\scriptsize Source domains: C, S, V.}
\vspace{-0.5em}
\begin{tabular}{|m{0.77cm}|m{0.25cm}m{0.75cm}m{0.86cm}|}
\hline
Domain & Base & M1 & M2 \\
\hline    
    L(New) & \textbf{60.73} & 59.89$\pm$0.1 & 60.73$\pm$0.2 \\  
    C($S_0$) & \textbf{96.70} & 94.81$\pm$0.2 & 96.30$\pm$0.5 \\  
    S($S_1$) & 75.91 & 78.23$\pm$0.3 & \textbf{78.32}$\pm$0.4 \\  
    V($S_2$) & \textbf{78.28} & 75.54$\pm$0.2 & 76.84$\pm$0.2 \\  
    Expanded & 77.91 & 77.12 & \textbf{78.05} \\ 
\hline
\end{tabular}
\end{subtable}
\\
\hspace{-1em}
\begin{subtable}{.492\linewidth}
\caption{\scriptsize Source domains: C, L, V.}
\vspace{-0.5em}
\begin{tabular}{|m{0.77cm}|m{0.25cm}m{0.75cm}m{0.86cm}|}
\hline
Domain & Base & M1 & M2 \\
\hline
    S(New) & 69.00 & \textbf{74.21}$\pm$0.2 & 72.92$\pm$0.7 \\  
    C($S_0$) & \textbf{98.11} & 98.04$\pm$0.2 & 97.88$\pm$0.4 \\  
    L($S_1$) & 64.99 & \textbf{65.32}$\pm$0.2 & 65.32$\pm$0.2  \\  
    V($S_2$) & 75.52 & \textbf{78.94}$\pm$0.2 & 78.27$\pm$0.2 \\  
    Expanded & 76.91 & \textbf{79.13} & 78.60 \\
\hline
\end{tabular}
\end{subtable}
\hspace{0.5em}
\begin{subtable}{.492\linewidth}
\caption{\scriptsize Source domains: C, S, L.}
\vspace{-0.5em}
\begin{tabular}{|m{0.77cm}|m{0.25cm}m{0.75cm}m{0.86cm}|}
\hline
Domain & Base & M1 & M2 \\
\hline
    V(New) & 68.90 & 68.75$\pm$0.1 & \textbf{69.67}$\pm$0.2 \\  
    C($S_0$) & \textbf{94.81} & 94.41$\pm$0.2 & 94.58$\pm$0.4 \\  
    S($S_1$) & 72.66 & \textbf{76.76}$\pm$0.2 & 76.86$\pm$0.2 \\  
    L($S_2$) & 66.62 & \textbf{69.44}$\pm$0.1 & 68.25$\pm$0.3 \\  
    Expanded & 75.74 & \textbf{77.34} & \textbf{77.34} \\
\hline
\end{tabular}
\end{subtable}
\end{center}
\end{scriptsize}
\vspace{-0.5em}
\end{table}

\begin{table}[!ht]
\begin{scriptsize}
\begin{center}
\caption{The accuracies (\%) on the imageCLEF\_DA dataset.}
\label{tab:imageCLEFUMSDE}
\vspace{-1em}
\hspace{-1em}
\begin{subtable}{.492\linewidth}
\caption{\scriptsize Source domains: C, I, P.}
\vspace{-0.5em}
\begin{tabular}{|m{0.77cm}|m{0.25cm}m{0.75cm}m{0.86cm}|}
\hline
Domain & Base & M1 & M2 \\
\hline    
    B(New) & \textbf{62.22} & 61.67$\pm$0.5 & 61.67$\pm$0.5 \\  
    C($S_0$) & \textbf{95.00} & 94.44$\pm$0 & \textbf{95.00}$\pm$0.5 \\  
    I($S_1$) & 90.56 & \textbf{91.72}$\pm$0.5 & 91.34$\pm$0.4  \\  
    P($S_2$) & 76.67 & \textbf{78.94}$\pm$0.4 & \textbf{78.89}$\pm$0 \\  
    Expanded & 81.11 & 81.69 & \textbf{81.72} \\
\hline
\end{tabular}
\end{subtable}
\hspace{0.5em}
\begin{subtable}{.492\linewidth}
\caption{\scriptsize Source domains: B, I, P.}
\vspace{-0.5em}
\begin{tabular}{|m{0.77cm}|m{0.25cm}m{0.75cm}m{0.86cm}|}
\hline
Domain & Base & M1 & M2 \\
\hline        
    C(New) & 93.33 & \textbf{94.39}$\pm$0.2 & 93.44$\pm$0.2 \\  
    B($S_0$) & 61.11 & \textbf{64.89}$\pm$0.5 & 63.89$\pm$0 \\  
    I($S_1$) & 90.56 & \textbf{92.33}$\pm$0.4 & 91.56$\pm$0.4 \\  
    P($S_2$) & 78.33 & \textbf{79.89}$\pm$0.2 & 79.39$\pm$0.2 \\  
    Expanded & 80.83 & \textbf{82.87} & 82.07\\
\hline
\end{tabular}
\end{subtable}
\\
\hspace{-1em}
\begin{subtable}{.492\linewidth}
\caption{\scriptsize Source domains: B, C, P.}
\vspace{-0.5em}
\begin{tabular}{|m{0.77cm}|m{0.25cm}m{0.75cm}m{0.86cm}|}
\hline
Domain & Base & M1 & M2 \\
\hline
    I(New) & 87.78 & \textbf{91.34}$\pm$0.7 & 90.78$\pm$0.5 \\  
    B($S_0$) & \textbf{67.78} & 63.83$\pm$0.4 & 65.28$\pm$0.5 \\  
    C($S_1$) & 95.00 & \textbf{96.06}$\pm$0.2 & 95.39$\pm$0.3  \\  
    P($S_2$) & 76.67 & \textbf{79.33}$\pm$0.2 & 78.05$\pm$0.5 \\  
    Expanded & 81.81 & \textbf{82.64} & 82.38 \\
\hline
\end{tabular}
\end{subtable}
\hspace{0.5em}
\begin{subtable}{.492\linewidth}
\caption{\scriptsize Source domains: B, I, C.}
\vspace{-0.5em}
\begin{tabular}{|m{0.77cm}|m{0.25cm}m{0.75cm}m{0.86cm}|}
\hline
Domain & Base & M1 & M2 \\
\hline
    P(New) & 76.11 & 76.22$\pm$0.2 & \textbf{77.17}$\pm$0.4 \\  
    B($S_0$) & \textbf{66.11} & 64.28$\pm$0.4 & 64.11$\pm$0.3 \\  
    I($S_1$) & \textbf{91.67} & 90.62$\pm$0.2 & 91.00$\pm$0.2 \\  
    C($S_2$) & 95.56 & 95.84$\pm$0.3 & \textbf{96.34}$\pm$0.4 \\  
    Expanded & \textbf{82.36} & 81.74 & 82.15  \\
\hline
\end{tabular}
\end{subtable}
\end{center}
\end{scriptsize}
\vspace{-0.5em}
\end{table}

\begin{table}[!ht]
\begin{scriptsize}
\begin{center}
\caption{The accuracies (\%) on the PACS dataset.}
\label{tab:PACSUMSDE}
\vspace{-1em}
\hspace{-1em}
\begin{subtable}{.492\linewidth}
\caption{\scriptsize Source domains: C, P, S.}
\vspace{-0.5em}
\begin{tabular}{|m{0.77cm}|m{0.25cm}m{0.75cm}m{0.86cm}|}
\hline
Domain & Base & M1 & M2 \\
\hline    
    A(New) & 79.80 & \textbf{86.50}$\pm$0.2 & 86.03$\pm$0.3 \\  
    C($S_0$) & 82.08 & \textbf{88.36}$\pm$0.3 & 87.61$\pm$0.4 \\  
    P($S_1$) & \textbf{96.81} & 95.41$\pm$0.1 & 95.83$\pm$0.2 \\  
    S($S_2$) & \textbf{94.83} & 86.28$\pm$0.5 & 87.80$\pm$0.4  \\  
    Expanded & 88.38 & 89.14 & \textbf{89.32} \\ 
\hline
\end{tabular}
\end{subtable}
\hspace{0.5em}
\begin{subtable}{.492\linewidth}
\caption{\scriptsize Source domains: A, P, S.}
\vspace{-0.5em}
\begin{tabular}{|m{0.77cm}|m{0.25cm}m{0.75cm}m{0.86cm}|}
\hline
Domain & Base & M1 & M2 \\
\hline        
    C(New) & 63.58 & \textbf{69.20}$\pm$0.4 & 67.92$\pm$0.2 \\  
    A($S_0$) & 86.97 & 89.37$\pm$0.3 & \textbf{90.05}$\pm$0.3 \\  
    P($S_1$) & \textbf{99.00} & 98.58$\pm$0.1 & 98.58$\pm$0.1 \\  
    S($S_2$) & 76.08 & \textbf{88.39}$\pm$0.2 & 87.19$\pm$0.2 \\  
    Expanded & 81.41 & \textbf{86.38} & 85.94 \\ 
\hline
\end{tabular}
\end{subtable}
\\
\hspace{-1em}
\begin{subtable}{.492\linewidth}
\caption{\scriptsize Source domains: A, C, S.}
\vspace{-0.5em}
\begin{tabular}{|m{0.77cm}|m{0.25cm}m{0.75cm}m{0.86cm}|}
\hline
Domain & Base & M1 & M2 \\
\hline
    P(New) & 96.01 & \textbf{97.72}$\pm$0.2 & 97.19$\pm$0.1 \\  
    A($S_0$) & 91.69 & \textbf{93.08}$\pm$0.3 & 93.08$\pm$0.3 \\  
    C($S_1$) & 82.50 & \textbf{85.82}$\pm$0.4 & 84.50$\pm$0.3 \\  
    S($S_2$) & \textbf{95.59} & 92.90$\pm$0.2 & 94.52$\pm$0.2 \\  
    Expanded & 91.45 & \textbf{92.45} & 92.32 \\ 
\hline
\end{tabular}
\end{subtable}
\hspace{0.5em}
\begin{subtable}{.492\linewidth}
\caption{\scriptsize Source domains: A, P, C.}
\vspace{-0.5em} 
\begin{tabular}{|m{0.77cm}|m{0.25cm}m{0.75cm}m{0.86cm}|}
\hline
Domain & Base & M1 & M2 \\
\hline
    S(New) & 61.75 & \textbf{74.52}$\pm$0.2 & 74.15$\pm$0.2 \\  
    A($S_0$) & \textbf{93.49} & 92.09$\pm$0.2 & 92.87$\pm$0.3 \\  
    P($S_1$) & \textbf{98.40} & 97.07$\pm$0.3 & 98.20$\pm$0 \\  
    C($S_2$) & 81.93 & \textbf{87.52}$\pm$0.5 & 85.61$\pm$0.2 \\  
    Expanded & 83.89 & \textbf{87.80} & 87.71 \\ 
\hline
\end{tabular}
\end{subtable}
\end{center}
\end{scriptsize}
\vspace{-0.5em}
\end{table}

\subsection{Results}
Tables~\ref{tab:VLCSUMSDE},~\ref{tab:imageCLEFUMSDE}, and~\ref{tab:PACSUMSDE} show the results from 
10 runs ($\pm$ standard deviation)
using the deep source models. 
The performance of the new domain test data (New), the multiple source domain test data ($S_0$, $S_1$, and $S_2$), as well as that on the target (i.e. expanded) domain calculated as the average over the new domain and the multiple source domains are presented. In the tables, {M1} and {M2} represent the fusion methods defined by Eqs.~(\ref{eqt:fusion1}) and~(\ref{eqt:fusion2}) respectively. The results are also compared with a baseline, denoted as ``Base'' in the tables. The baseline is defined as the sum fusion of the original source models. 
Note that since the source data are not available (which is unlike the cases in domain adaptation) and the UMSDE is concerned with the performance on both source and new domains, it is not possible and probably unfair to compare the proposed unsupervised UMSDE method with a DA method that requires source data.

As shown in the tables, the proposed methods outperformed the baseline in the target or expanded domain in eleven of the twelve cases over the three datasets. The only exceptional case is on the imageCLEF\_DA dataset in which the proposed method performed marginally worse (i.e. less than one percentage point). The two fusion methods (i.e. $M1$ and $M2$ in the tables) performed comparably. 
In addition, the standard deviation of different repeats of each experiment is small, indicating that the proposed methods are robust against different runs.

Considering the performance in the new domain, the proposed method outperformed the baseline in most of the cases. In six of the twelve cases, it is by a large margin. Notice that there are two cases that the proposed method did not improve the performance in the new domain compared to the baseline. This is probably that the distributions of the multiple source domains overlap with that of the new domain, or the distribution shift cannot be characterized merely by the source models. Comparing the two fusion methods, it can be seen that $M1$ generally outperforms $M2$ on the new domain data as expected since only the models updated by the new domain are used in $M1$. 


Also as expected, the performance improvement in the new domain does not degrade the performance in the source domains in most of the cases. In contrast, the performances in the source domains are even improved after expansion using the proposed method in more than half of the cases over the three datasets. This is probably because the learned model not only deals with the bias between the source domains and the new domain, but also deals with the bias among multiple source domains.

Experiments were conducted using max fusion over the original models and over the updated models, the improvement of the performance over the various cases have the similar trend as using the sum fusion. Therefore, only results of sum fusion are presented in the paper.

\subsection{Discussion and Analysis}

\subsubsection{Performance of Updated Source Models}
To show how well the individual updated source models by the proposed method work, experiments were conducted to compare the performance of the original source models and the updated models in every source domain. Table~\ref{tab:VLCSsingle} shows the results of the case (L,S,V$\rightarrow$C) on the VLSC dataset using deep source models. It can be seen that after expansion using the proposed method, the performance of each updated source model is not only improved in the new domain but also improved in other source domains. This verifies that the updated model not only deals with the bias between source domains and the new domain, but also deals with the bias among source domains. This trend is also observed in other cases on the other two datasets.

\begin{table}[!ht]
\begin{scriptsize}
\begin{center}
\caption{The accuracies (\%) in individual domains using individual original and updated source models for the case of  L,S,V$\rightarrow$C on the VLCS dataset where ``Orig." refers to the original models and ``Upd." represents the updated models.}
\vspace{-0.5em}
\begin{tabu}{|m{0.65cm}|m{0.4cm}m{0.5cm}|[1.5pt]m{0.65cm}|m{0.4cm}m{0.5cm}|[1.5pt]m{0.65cm}|m{0.4cm}m{0.5cm}|}
\hline
Domain & Orig. & Upd. & Domain & Orig. & Upd. & Domain & Orig. & Upd. \\
\hline
C & 85.61 & \textbf{93.40} & C & 43.63 & \textbf{95.05} & C & 95.28 & \textbf{95.99}\\
L & \textbf{75.91} & 75.41 & L & \textbf{58.85} & 58.59 & L & 57.97 & \textbf{58.09}\\
S & 49.29 & \textbf{52.95} & S & \textbf{79.57} & 74.39 &S & 71.85 & \textbf{72.15}\\
V & 58.93 & \textbf{63.97} & V & 58.34 & \textbf{62.88} & V & \textbf{80.55} & 80.36\\
Average & 67.44 & \textbf{71.43} & Average & 60.10 & \textbf{72.73} & Average & 76.41 & \textbf{76.65}\\
\hline
\multicolumn{3}{|c|[1.5pt]}{Source L model} & \multicolumn{3}{c|[1.5pt]}{Source S model} & \multicolumn{3}{c|}{Source V model} \\
\hline
\end{tabu}
\label{tab:VLCSsingle}
\end{center}
\end{scriptsize}
\vspace{-0.5em}
\end{table}

\subsubsection{Performance using Shallow Models}
To further demonstrate the robustness of the proposed method, experiments were conducted using shallow source models which are a  multi-layer perceptron with one-layer of 1000 hidden neurons on the DECAF~\cite{Donahue2014} features. Results are shown in Table~\ref{tab:MSDAresults}. It can be seen that the proposed method also works well on the shallow source models and achieved better results than the baseline method.

\begin{table}[!ht]
\begin{scriptsize}
\begin{center}
\caption{The average accuracies (\%) obtained on the expanded domain using shallow models on the three datasets and the average accuracies (\%) over all the datasets.}
\vspace{-0.5em}
\begin{tabu}{|m{1.5cm}|m{0.7cm}m{0.7cm}m{0.7cm}|}
\hline
Datasets & Base & M1 & M2 \\
\hline
VLCS & 73.69 & \textbf{75.82} & 75.51 \\
imageCLEF & 77.67 & \textbf{78.47} & 78.44 \\
PACS & 77.11 & 76.12 & \textbf{77.21} \\
\hline
Average & 76.16 & 76.81 & \textbf{77.05} \\
\hline
\end{tabu}
\label{tab:MSDAresults}
\end{center}
\end{scriptsize}
\vspace{-1.5em}
\end{table}

\subsection{Ablation Studies and Parameter Sensitivities}
We conducted ablation experiments to validate the effectiveness of the proposed method for learning the source model weights and evaluate the sensitivities of parameters.
\vspace{-0.5em}
\subsubsection{On the Source Model Importance Weights} 
\begin{figure}[!ht]
\begin{center}
\includegraphics[scale=0.6]{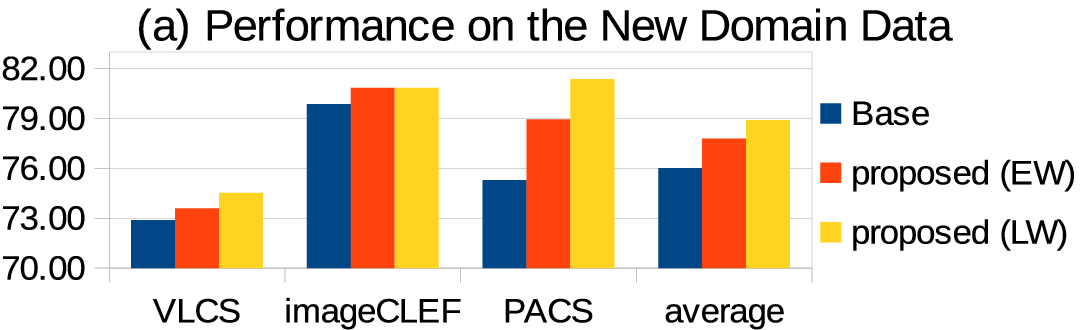}
\includegraphics[scale=0.6]{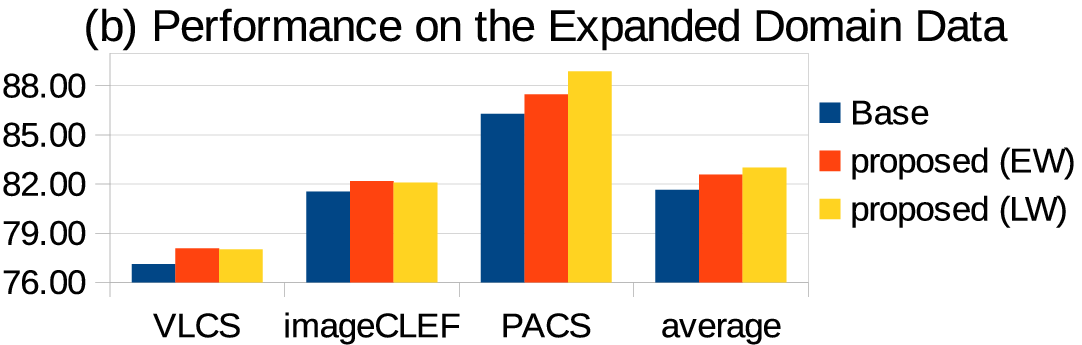}
\caption{Comparison of the performance on the new and expanded domain data using equal weights (EW) and using the proposed learned weights (LW).
}
\label{fig:weights}
\end{center}
\vspace{-0.5em}
\end{figure}

The first ablation study is to evaluate the effectiveness of the proposed source model weights when aligning the multiple source domains and the new domain. 
We compared the cases with equal weights and learned weights in Eq.~\ref{eqt:overall}.
Results for the new domain, expanded domain and the average over the two domains on the three datasets are shown in Figure~\ref{fig:weights}. It can be seen that when the source models are assigned with equal weights, the proposed method can still obtain better results compared to the baselines and using the learned source model weights can further improve the results on average on both the new domain and the expanded domain. 
Table~\ref{tab:VLCSweightsvalue} shows the accuracies obtained on the new domain using a single original source model and the learned weights of each source model using Eq.~\ref{eqt:weights}.
It can be seen that the higher accuracy in a specific new domain generally corresponds to a lower weight while the lower accuracy generally corresponds to a higher weight. This is consistent with our assumption that the better the performance the smaller the entropy.

\begin{table}[!ht]
\begin{scriptsize}
\begin{center}
\vspace{-1.5em}
\caption{The the accuracies (\%) obtained on the new domain using a single original source model and the learned weight of each source model using the proposed method.}
\vspace{-1em}
\begin{tabu}{|m{0.5cm}|m{0.45cm}m{0.5cm}|[1.5pt]m{0.5cm}|m{0.45cm}m{0.5cm}|[1.5pt]m{0.5cm}|m{0.45cm}m{0.5cm}|}
\hline
\multirow{2}{1.cm}{Model} & \multicolumn{2}{c|[1.5pt]}{VLCS} &\multirow{2}{1.cm}{Model} & \multicolumn{2}{c|[1.5pt]}{imageCLEF} & \multirow{2}{1.cm}{Model} &\multicolumn{2}{c|}{PACS} \\
\cline{2-3} 
\cline{5-6} 
\cline{8-9} 
  & Acc. & weight & & Acc. & weight & & Acc. & weight \\
\hline
$L$ & 85.61 & 0.1139 & $C$ & 58.89 & 0.54 & $C$ & 78.50 & 0.2304 \\
$S$ & 43.63 & 0.8476 & $I$ & 56.67 & 0.2247 & $P$ & 66.12 & 0.3513 \\
$V$ & 95.28 & 0.0385 & $P$ & 54.44 & 0.2353 & $S$& 58.14 & 0.4183 \\
\hline
\multicolumn{3}{|c|[1.5pt]}{New Domain: $C$ } & \multicolumn{3}{c|[1.5pt]}{New Domain: $B$ } & \multicolumn{3}{c|}{New Domain: $A$ } \\
\hline
\hline
$C$ & 55.96 & 0.7104 & $B$ & 93.33 & 0.3247 & $A$ & 62.73 & 0.1856 \\
$S$ & 58.85 & 0.1451 & $I$ & 88.89 & 0.2715 & $P$ & 35.99 & 0.6870 \\
$V$ & 57.97 & 0.1445 & $P$ & 88.33 & 0.4038 & $S$ & 58.61 & 0.1273 \\
\hline
\multicolumn{3}{|c|[1.5pt]}{New Domain: $L$ } & \multicolumn{3}{c|[1.5pt]}{New Domain: $C$ } & \multicolumn{3}{c|}{New Domain: $C$ } \\
\hline
\hline
$C$ & 52.74 & 0.5748 & $B$ & 80.56 & 0.3578 & $A$ & 96.01 & 0.0703 \\
$L$ & 49.29 & 0.2795 & $C$ & 77.22 & 0.3927 & $C$ & 82.83 & 0.1525 \\
$V$ & 71.85 & 0.1457 & $P$ & 85.56 & 0.2495 & $S$ & 60.08 & 0.7773 \\
\hline
\multicolumn{3}{|c|[1.5pt]}{New Domain: $S$ } & \multicolumn{3}{c|[1.5pt]}{New Domain: $I$ } & \multicolumn{3}{c|}{New Domain: $P$ } \\
\hline
\hline
$C$ & 55.97 & 0.4052 & $B$ & 71.76 & 0.3205 & $A$ & 52.25 & 0.1441 \\
$S$ & 58.34 & 0.1957 & $I$ & 77.22 & 0.1946 & $P$ & 27.57 & 0.7589 \\
$L$ & 58.93 & 0.3991 & $C$ & 67.22 & 0.485 & $C$ & 62.68 & 0.097 \\
\hline
\multicolumn{3}{|c|[1.5pt]}{New Domain: $V$ } & \multicolumn{3}{c|[1.5pt]}{New Domain: $P$ } & \multicolumn{3}{c|}{New Domain: $S$ } \\
\hline
\end{tabu}
\label{tab:VLCSweightsvalue}
\end{center}
\end{scriptsize}
\vspace{-1.5em}
\end{table}    

\vspace{-1em}
\subsubsection{On the Temperature}
\begin{figure}[!h]

\minipage{0.23\textwidth}
\centering
  \includegraphics[scale=0.5]{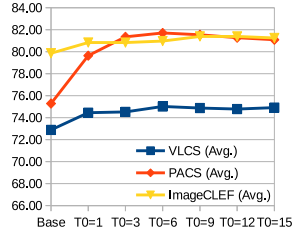}
  \vspace{-1em}
  \caption{\small The accuracy when changing the temperature.}
  \label{fig:temperature}
\endminipage\hfill
\minipage{0.23\textwidth}
\centering
  \includegraphics[scale=0.5]{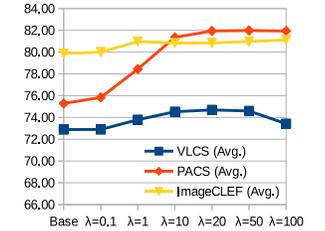}
  \vspace{-1em}
  \caption{\small The accuracy when changing $\lambda$.}
  \label{fig:lambda}
\endminipage
\vspace{-0.5em}
\end{figure}

The second ablation study is to validate the effectiveness of the use of a proper temperature when aligning different domains as well as adapting the source information to the new domain. Experiments were conducted on the three datasets and the average accuracies on the new domain are reported because the performance of the source domain is not sensitive to the temperature. Figure~\ref{fig:temperature} shows the average accuracy over the four cases on each dataset when changing the value of the temperature. It shows that when the temperature increases, the improvement of performance can be observed on most of the datasets. However, the performance will drop slightly if the temperature is too large because the class relationship information would be destroyed since the probabilities of all classes would be similar. In addition, even with a small temperature of $1.0$, the proposed method outperforms the baseline. It indicates that the proposed method can effectively adapt multiple source models to the new domain. In addition, a higher temperature value produces a softer probability distribution over classes, and thus more information on class relationships can be preserved, resulting in the improvement of performances.

\vspace{-1.5em}
\subsubsection{On the Trade-off Parameter} 
This study aims at evaluating the sensitivity of the trade-off parameter $\lambda$. 
The accuracies on the new domain are reported since the results of the source domain test data are not sensitive to $\lambda$. 
Figure~\ref{fig:lambda} demonstrates the average accuracy over four cases on each dataset when changing the value of $\lambda$ and shows that when the $\lambda$ getting large, the performances on the new domain are improved because the bias among domains would be largely reduced. If the $\lambda$ is too large (i.e. $\lambda=100$), the performance in the source domains could be adversely affected but can be alleviated by including the original source models in the fusion (e.g. M2). Hence, it can be seen that there is a large range of $\lambda$ values in which the performance can be improved compared to the baseline, therefore, $\lambda$ is easy to choose.

\section{Conclusion}
This paper is concerned with unsupervised multi-source domain expansion (UMSDE), where the target domain is formed jointly by the source domains and new domain. Thus the learned label function is expected to work equally well for all source domains and the new domain. Specifically, this paper proposes a method for an unsupervised UMSDE problem where only models of source domains and unlabelled new domain data are available. A possible venue to further improve the proposed method is to effectively fuse the original source models and updated models.

The proposed concept of Unsupervised Multi-source Domain expansion (UMSDE) captures a new research dimension in transfer learning. Though keeping the performance in the source domains has sometimes been used to as constraints in traditional domain adaptation (DA)~\cite{Zhang2019}, this has never formally been a compulsory requirement in DA. In addition, UMSDE has the potential to formally bridge DA with online and incremental learning where the target domain is continuously expanded. Many problems in domain expansion are yet to be investigated in the future.

{\small
\bibliographystyle{IEEEtran} 
\bibliography{CrossDataset}

\begin{thebibliography}{10}
\providecommand{\url}[1]{#1}
\csname url@samestyle\endcsname
\providecommand{\newblock}{\relax}
\providecommand{\bibinfo}[2]{#2}
\providecommand{\BIBentrySTDinterwordspacing}{\spaceskip=0pt\relax}
\providecommand{\BIBentryALTinterwordstretchfactor}{4}
\providecommand{\BIBentryALTinterwordspacing}{\spaceskip=\fontdimen2\font plus
\BIBentryALTinterwordstretchfactor\fontdimen3\font minus
  \fontdimen4\font\relax}
\providecommand{\BIBforeignlanguage}[2]{{%
\expandafter\ifx\csname l@#1\endcsname\relax
\typeout{** WARNING: IEEEtran.bst: No hyphenation pattern has been}%
\typeout{** loaded for the language `#1'. Using the pattern for}%
\typeout{** the default language instead.}%
\else
\language=\csname l@#1\endcsname
\fi
#2}}
\providecommand{\BIBdecl}{\relax}
\BIBdecl

\bibitem{Krizhevsky2012}
A.~Krizhevsky, I.~Sutskever, and G.~E. Hinton, ``Imagenet classification with
  deep convolutional neural networks,'' in \emph{Advances in neural information
  processing systems}, 2012, pp. 1097--1105.

\bibitem{Zhang2019}
J.~Zhang, W.~Li, P.~Ogunbona, and D.~Xu, ``Recent advances in transfer learning
  for cross-dataset visual recognition: A problem-oriented perspective,''
  \emph{ACM Computing Surveys (CSUR)}, vol.~52, no.~1, p.~7, 2019.

\bibitem{Shao2015}
L.~Shao, F.~Zhu, and X.~Li, ``Transfer learning for visual categorization: a
  survey,'' \emph{IEEE Transactions on Neural Networks and Larning Systems},
  vol.~26, no.~5, pp. 1019--1034, 2015.

\bibitem{Ben-David2010}
S.~Ben-David, J.~Blitzer, K.~Crammer, A.~Kulesza, F.~Pereira, and J.~W.
  Vaughan, ``A theory of learning from different domains,'' \emph{Machine
  learning}, vol.~79, no.~1, pp. 151--175, 2010.

\bibitem{Ganin2016}
Y.~Ganin, E.~Ustinova, H.~Ajakan, P.~Germain, H.~Larochelle, F.~Laviolette,
  M.~Marchand, and V.~Lempitsky, ``Domain-adversarial training of neural
  networks,'' \emph{Journal of Machine Learning Research}, vol.~17, no.~59, pp.
  1--35, 2016.

\bibitem{Long2017}
M.~Long, J.~Wang, and M.~I. Jordan, ``Deep transfer learning with joint
  adaptation networks,'' in \emph{Proc. International Conference on Machine
  Learning}, 2017.

\bibitem{Chidlovskii2016}
B.~Chidlovskii, S.~Clinchant, and G.~Csurka, ``Domain adaptation in the absence
  of source domain data,'' in \emph{Proc. ACM SIGKDD International Conference
  on Knowledge Discovery and Data Mining}.\hskip 1em plus 0.5em minus
  0.4em\relax ACM, 2016, pp. 451--460.

\bibitem{Khosla2012}
A.~Khosla, T.~Zhou, T.~Malisiewicz, A.~A. Efros, and A.~Torralba, ``Undoing the
  damage of dataset bias,'' in \emph{Proc. European Conference on Computer
  Vision}.\hskip 1em plus 0.5em minus 0.4em\relax Springer, 2012, pp. 158--171.

\bibitem{Li2017d}
Z.~Li and D.~Hoiem, ``Learning without forgetting,'' \emph{IEEE Transactions on
  Pattern Analysis and Machine Intelligence}, 2017.

\bibitem{Jung2018}
H.~Jung, J.~Ju, M.~Jung, and J.~Kim, ``Less-forgetful learning for domain
  expansion in deep neural networks,'' in \emph{Proc. Thirty-Second AAAI
  Conference on Artificial Intelligence}, 2018.

\bibitem{Nelakurthi2018}
A.~R. Nelakurthi, R.~Maciejewski, and J.~He, ``Source free domain adaptation
  using an off-the-shelf classifier,'' in \emph{Proc. IEEE International
  Conference on Big Data}.\hskip 1em plus 0.5em minus 0.4em\relax IEEE, 2018,
  pp. 140--145.

\bibitem{AbdullahJamal2018}
M.~Abdullah~Jamal, H.~Li, and B.~Gong, ``Deep face detector adaptation without
  negative transfer or catastrophic forgetting,'' in \emph{Proc. IEEE
  Conference on Computer Vision and Pattern Recognition}, 2018, pp. 5608--5618.

\bibitem{Li2017b}
D.~Li, Y.~Yang, Y.-Z. Song, and T.~M. Hospedales, ``Deeper, broader and artier
  domain generalization,'' in \emph{Proc. IEEE Conference on Computer Vision
  and Pattern Recognition}, 2017, pp. 5542--5550.

\bibitem{Lee2017}
S.-W. Lee, J.-H. Kim, J.~Jun, J.-W. Ha, and B.-T. Zhang, ``Overcoming
  catastrophic forgetting by incremental moment matching,'' in \emph{Advances
  in Neural Information Processing Systems}, 2017, pp. 4655--4665.

\bibitem{Zenke2017}
F.~Zenke, B.~Poole, and S.~Ganguli, ``Continual learning through synaptic
  intelligence,'' in \emph{International Conference on Machine Learning}, 2017,
  pp. 3987--3995.

\bibitem{Pan2010}
S.~J. Pan and Q.~Yang, ``A survey on transfer learning,'' \emph{IEEE
  Transactions on Knowledge and Data Engineering}, vol.~22, no.~10, pp.
  1345--1359, 2010.

\bibitem{Pan2011}
S.~J. Pan, I.~W. Tsang, J.~T. Kwok, and Q.~Yang, ``Domain adaptation via
  transfer component analysis,'' \emph{IEEE Transactions on Neural Networks},
  vol.~22, no.~2, pp. 199--210, 2011.

\bibitem{Zhang2017}
J.~Zhang, W.~Li, and P.~Ogunbona, ``Joint geometrical and statistical alignment
  for visual domain adaptation,'' in \emph{Proc. IEEE Conference on Computer
  Vision and Pattern Recognition}, 2017.

\bibitem{Long2015a}
M.~Long and J.~Wang, ``Learning transferable features with deep adaptation
  networks,'' in \emph{Proc. International Conference on Machine Learning},
  2015, pp. 97--105.

\bibitem{Tzeng2017}
E.~Tzeng, J.~Hoffman, K.~Saenko, and T.~Darrell, ``Adversarial discriminative
  domain adaptation,'' \emph{Proc. IEEE Conference on Computer Vision and
  Pattern Recognition}, 2017.

\bibitem{Saito2018}
K.~Saito, K.~Watanabe, Y.~Ushiku, and T.~Harada, ``Maximum classifier
  discrepancy for unsupervised domain adaptation,'' in \emph{Proc. IEEE
  International Conference on Computer Vision}, 2018.

\bibitem{Sun2015a}
S.~Sun, H.~Shi, and Y.~Wu, ``A survey of multi-source domain adaptation,''
  \emph{Information Fusion}, vol.~24, pp. 84--92, 2015.

\bibitem{Erfani2017}
S.~M. Erfani, M.~Baktashmotlagh, M.~Moshtaghi, V.~Nguyen, C.~Leckie, J.~Bailey,
  and K.~Ramamohanarao, ``From shared subspaces to shared landmarks: A robust
  multi-source classification approach.'' in \emph{Proc. AAAI Conference on
  Artificial Intelligence}, 2017.

\bibitem{Xu2018}
R.~Xu, Z.~Chen, W.~Zuo, J.~Yan, and L.~Lin, ``Deep cocktail network:
  Multi-source unsupervised domain adaptation with category shift,'' in
  \emph{Proc. IEEE Conference on Computer Vision and Pattern
  Recognition}.\hskip 1em plus 0.5em minus 0.4em\relax IEEE, 2018.

\bibitem{Blanchard2011}
G.~Blanchard, G.~Lee, and C.~Scott, ``Generalizing from several related
  classification tasks to a new unlabeled sample,'' in \emph{Proc. Advances in
  Neural Information Processing Systems}, 2011, pp. 2178--2186.

\bibitem{Muandet2013}
K.~Muandet, D.~Balduzzi, and B.~Sch{\"o}lkopf, ``Domain generalization via
  invariant feature representation,'' in \emph{Proc. International Conference
  on Machine Learning}, 2013, pp. 10--18.

\bibitem{Xu2014}
Z.~Xu, W.~Li, L.~Niu, and D.~Xu, ``Exploiting low-rank structure from latent
  domains for domain generalization,'' in \emph{Proc. European Conference on
  Computer Vision}.\hskip 1em plus 0.5em minus 0.4em\relax Springer, 2014, pp.
  628--643.

\bibitem{Hinton2014}
G.~Hinton, O.~Vinyals, and J.~Dean, ``Distilling the knowledge in a neural
  network,'' in \emph{NIPS 2014 Deep Learning Workshop}, 2014.

\bibitem{Grandvalet2005}
Y.~Grandvalet and Y.~Bengio, ``Semi-supervised learning by entropy
  minimization,'' in \emph{Advances in neural information processing systems},
  2005, pp. 529--536.

\bibitem{Mansour2009}
Y.~Mansour, M.~Mohri, and A.~Rostamizadeh, ``Domain adaptation with multiple
  sources,'' in \emph{Advances in neural information processing systems}, 2009,
  pp. 1041--1048.

\bibitem{Duan2012c}
L.~Duan, D.~Xu, and I.~W. Tsang, ``Domain adaptation from multiple sources: A
  domain-dependent regularization approach,'' \emph{IEEE Transactions on Neural
  Networks and Learning Systems}, vol.~23, no.~3, pp. 504--518, 2012.

\bibitem{Torralba2011}
A.~Torralba and A.~Efros, ``Unbiased look at dataset bias,'' in \emph{Proc.
  IEEE Conference on Computer Vision and Pattern Recognition}.\hskip 1em plus
  0.5em minus 0.4em\relax IEEE, 2011, pp. 1521--1528.

\bibitem{Everingham}
M.~Everingham, L.~Van~Gool, C.~K.~I. Williams, J.~Winn, and A.~Zisserman, ``The
  {PASCAL} {V}isual {O}bject {C}lasses {C}hallenge 2007 {(VOC2007)}
  {R}esults.''

\bibitem{Russell2008}
B.~C. Russell, A.~Torralba, K.~P. Murphy, and W.~T. Freeman, ``Labelme: a
  database and web-based tool for image annotation,'' \emph{International
  journal of computer vision}, vol.~77, no. 1-3, pp. 157--173, 2008.

\bibitem{Fei-Fei2006}
L.~Fei-Fei, R.~Fergus, and P.~Perona, ``One-shot learning of object
  categories,'' \emph{IEEE Transactions on Pattern Analysis and Machine
  Intelligence}, vol.~28, no.~4, pp. 594--611, 2006.

\bibitem{Choi2010}
M.~J. Choi, J.~J. Lim, A.~Torralba, and A.~S. Willsky, ``Exploiting
  hierarchical context on a large database of object categories,'' in
  \emph{Proc. IEEE Conference on Computer VIsion and Pattern Recognition},
  2010.

\bibitem{Griffin2007}
G.~Griffin, A.~Holub, and P.~Perona, ``Caltech-256 object category dataset,''
  Tech. Rep., 2007.

\bibitem{Deng2009}
J.~Deng, W.~Dong, R.~Socher, L.-J. Li, K.~Li, and L.~Fei-Fei, ``Imagenet: A
  large-scale hierarchical image database,'' in \emph{Proc. IEEE Conference on
  Computer Vision and Pattern Recognition}.\hskip 1em plus 0.5em minus
  0.4em\relax IEEE, 2009, pp. 248--255.

\bibitem{Bergamo2010}
A.~Bergamo and L.~Torresani, ``Exploiting weakly-labeled web images to improve
  object classification: a domain adaptation approach,'' in \emph{Advances in
  neural information processing systems}, 2010, pp. 181--189.

\bibitem{Sangkloy2016}
P.~Sangkloy, N.~Burnell, C.~Ham, and J.~Hays, ``The sketchy database: learning
  to retrieve badly drawn bunnies,'' \emph{ACM Transactions on Graphics (TOG)},
  vol.~35, no.~4, p. 119, 2016.

\bibitem{Eitz2012}
M.~Eitz, J.~Hays, and M.~Alexa, ``How do humans sketch objects?'' \emph{ACM
  Transactions on Graphics}, vol.~31, no.~4, p.~44, 2012.

\bibitem{He2016}
K.~He, X.~Zhang, S.~Ren, and J.~Sun, ``Deep residual learning for image
  recognition,'' in \emph{Proc. Conference on Computer Vision and Pattern
  Recognition}, 2016, pp. 770--778.

\bibitem{Jia2014a}
Y.~Jia, E.~Shelhamer, J.~Donahue, S.~Karayev, J.~Long, R.~Girshick,
  S.~Guadarrama, and T.~Darrell, ``Caffe: Convolutional architecture for fast
  feature embedding,'' in \emph{Proceedings of the 22nd ACM international
  conference on Multimedia}.\hskip 1em plus 0.5em minus 0.4em\relax ACM, 2014,
  pp. 675--678.

\bibitem{Donahue2014}
J.~Donahue, Y.~Jia, O.~Vinyals, J.~Hoffman, N.~Zhang, E.~Tzeng, and T.~Darrell,
  ``Decaf: A deep convolutional activation feature for generic visual
  recognition,'' in \emph{Proc. International Conference on Machine Learning},
  2014, pp. 647--655.

\end{thebibliography}
}

\end{document}